\documentclass[journal]{IEEEtran}

\usepackage{amsmath,amsfonts}
\usepackage{algorithmic}
\usepackage{algorithm}
\usepackage[caption=false,font=normalsize,labelfont=sf,textfont=sf]{subfig}
\usepackage{textcomp}
\usepackage{url}
\usepackage{verbatim}
\usepackage{graphicx}
\usepackage{cite}
\usepackage{xcolor}
\usepackage{balance} 
\usepackage{silence} 

\usepackage[
    colorlinks=true,
    linkcolor=blue,
    citecolor=blue,
    urlcolor=black
]{hyperref}

\usepackage{orcidlink} 

\usepackage{etoolbox}
\AtBeginDocument{%
    \renewcommand{\cite}[1]{\textcolor{blue}{[\citenum{#1}]}}
}

\usepackage{csquotes}      

\usepackage[capitalise]{cleveref}	
\crefname{figure}{Fig.}{Figs.} 
\Crefname{figure}{Fig.}{Figs.} 
\crefname{equation}{Eq.}{Eqs.}
\Crefname{equation}{Eq.}{Eqs.}

\usepackage{color,soul}
\usepackage{amssymb}   
\usepackage{pifont}         
\newcommand{\cmark}{\ding{51}}  
\newcommand{\xmark}{\ding{55}}  
\newcommand{\dmark}{\ding{108}}

\usepackage{array}
\usepackage{booktabs}
\usepackage{vcell}
\usepackage{tabularx} 
\usepackage{fontsize} 
\usepackage{makecell}
\usepackage{multirow}  
\usepackage{adjustbox} 

\definecolor{mydarkgreen}{RGB}{0,200,0}

\usepackage{tabularray}
\usepackage{acro}[=v2]   
\usepackage{enumitem}
\usepackage{mfirstuc}  
\usepackage{microtype}  

\setlist{nolistsep}    

\DeclareAcroListStyle{mydesc}{list}{
  list = description ,
}

\acsetup{
  list-style = mydesc
}

\setlist[description]{style=standard, leftmargin=2.5cm, labelsep=0pt, labelwidth=\dimexpr2.5cm-\labelsep}

\DeclareAcronym{AI}{
	short = AI,
	long  = artificial intelligence,
	class = abbrev
}

\DeclareAcronym{GenAI}{
	short = GenAI,
	long  = generative artificial intelligence,
	class = abbrev
}

\DeclareAcronym{GenAIs}{
	short = GenAIs,
	long  = generative artificial intelligences,
	class = abbrev
}

\DeclareAcronym{DAI}{
	short = DAI,
	long  = discriminative artificial intelligence,
	class = abbrev
}

\DeclareAcronym{ML}{
	short = ML,
	long  = machine learning,
	class = abbrev
}

\DeclareAcronym{IAI}{
	short = IAI,
	long  = interactive AI,
	class = abbrev
}

\DeclareAcronym{PV}{
	short = PV,
	long  = photovoltaic,
	class = abbrev
}

\DeclareAcronym{CSOs}{
	short = CSOs,
	long  = charging station operators,
	class = abbrev
}

\DeclareAcronym{EV}{
	short = EV,
	long  = electric vehicle,
	class = abbrev
}

\DeclareAcronym{LLMs}{
	short = LLMs,
	long  = large language models,
	class = abbrev
}

\DeclareAcronym{LLM}{
	short = LLM,
	long  = large language model,
	class = abbrev
}

\DeclareAcronym{RAG}{
	short = RAG,
	long  = retrieval-augmented generation,
	class = abbrev
}

\DeclareAcronym{SoC}{
	short = SoC,
	long  = state-of-charge,
	class = abbrev
}

\DeclareAcronym{NLP}{
	short = NLP,
	long  = natural language processing,
	class = abbrev
}

\DeclareAcronym{DRL}{
	short = DRL,
	long  = deep reinforcement learning,
	class = abbrev
}

\DeclareAcronym{RL}{
	short = RL,
	long  = reinforcement learning,
	class = abbrev
}

\DeclareAcronym{AC}{
	short = AC,
	long  = alternating current,
	class = abbrev
}

\DeclareAcronym{DC}{
	short = DC,
	long  = direct current,
	class = abbrev
}

\DeclareAcronym{TSP}{
	short = TSP,
	long  = traveling salesman problem,
	class = abbrev
}

\DeclareAcronym{BESS}{
	short = BESS,
	long  = battery energy storage system,
	class = abbrev
}

\DeclareAcronym{FEWL}{
	short = FEWL,
	long  = factualness evaluations via weighting LLMs,
	class = abbrev
}

\DeclareAcronym{BERT}{
	short = BERT,
	long  = bidirectional encoder representations from Transformers,
	class = abbrev
}

\DeclareAcronym{GPT}{
	short = GPT,
	long  = generative pre-trained Transformer,
	class = abbrev
}

\DeclareAcronym{XLNet}{
	short = XLNet,
	long  = eXtreme multi-label text classification with Transformers,
	class = abbrev
}

\DeclareAcronym{T5}{
	short = T5,
	long  = text-to-text transfer Transformer,
	class = abbrev
}

\DeclareAcronym{GRU}{
	short = GRU,
	long  = gated recurrent unit,
	class = abbrev
}

\DeclareAcronym{LSTM}{
	short = LSTM,
	long  = long short-term memory,
	class = abbrev
}

\DeclareAcronym{HEMS}{
	short = HEMS,
	long  = home energy management system,
	class = abbrev
}

\DeclareAcronym{EMS}{
	short = EMS,
	long  = energy management system,
	class = abbrev
}

\DeclareAcronym{TOU}{
	short = TOU,
	long  = time-of-use,
	class = abbrev
}

\DeclareAcronym{DAP}{
	short = DAP,
	long  = day-ahead pricing,
	class = abbrev
}

\DeclareAcronym{RTP}{
	short = RTP,
	long  = real-time pricing,
	class = abbrev
}

\DeclareAcronym{DSM}{
	short = DSM,
	long  = demand side management,
	class = abbrev
}

\DeclareAcronym{IoEV}{
	short = IoEV,
	long  = Internet of electric vehicles,
	class = abbrev
}

\DeclareAcronym{MILP}{
	short = MILP,
	long  = mixed integer linear programming,
	class = abbrev
}

\DeclareAcronym{DEG}{
	short = DEG,
	long  = diesel engine generator,
	class = abbrev
}

\DeclareAcronym{EWH}{
	short = EWH,
	long  = electric water heater,
	class = abbrev
}

\DeclareAcronym{HVAC}{
	short = HVAC,
	long  = heating-ventilation-air conditioning,
	class = abbrev
}

\DeclareAcronym{IBR}{
	short = IBR,
	long  = inclining block rate,
	class = abbrev
}

\DeclareAcronym{PAR}{
	short = PAR,
	long  = peak-to-average ratio,
	class = abbrev
}

\DeclareAcronym{DI}{
	short = DI,
	long  = discomfort index,
	class = abbrev
}

\DeclareAcronym{V2H}{
	short = V2H,
	long  = vehicle-to-home,
	class = abbrev
}

\DeclareAcronym{DR}{
	short = DR,
	long  = demand response,
	class = abbrev
}

\DeclareAcronym{GAN}{
	short = GAN,
	long  = generative adversarial network,
	class = abbrev 
}

\DeclareAcronym{GMM}{
	short = GMM,
	long  = Gaussian mixture model,
	class = abbrev
}

\DeclareAcronym{GDM}{
	short = GDM,
	long  = generative diffusion model,
	class = abbrev
}

\DeclareAcronym{VAE}{
	short = VAE,
	long  = variational autoencoder,
	class = abbrev
}

\DeclareAcronym{DNN}{
	short = DNN,
	long  = deep neural network,
	class = abbrev
}

\DeclareAcronym{CopulaGAN}{
	short = CopulaGAN,
	long  = Copula generative adversarial network,
	class = abbrev
}

\DeclareAcronym{DBSCAN}{
	short = DBSCAN,
	long  = density-based spatial clustering of applications with noise,
	class = abbrev
}

\DeclareAcronym{ESS}{
	short = ESS,
	long  = energy storage system,
	class = abbrev
}

\DeclareAcronym{GA}{
	short = GA,
	long  = genetic algorithms,
	class = abbrev
}

\DeclareAcronym{DP}{
	short = DP,
	long  = dynamic programming,
	class = abbrev
}

\DeclareAcronym{PSO}{
	short = PSO,
	long  = particle swarm optimization,
	class = abbrev
}

\DeclareAcronym{SO}{
	short = SO,
	long  = stochastic optimization,
	class = abbrev
}

\DeclareAcronym{EMA}{
	short = EMA,
	long  = Energy Market Authority ,
	class = abbrev
}

\DeclareAcronym{RNN}{
	short = RNN,
	long  = recurrent neural network,
	class = abbrev
}

\DeclareAcronym{ES}{
	short = ES,
	long  = energy storage,
	class = abbrev
}

\DeclareAcronym{GANs}{
	short = GANs,
	long  = generative adversarial networks,
	class = excluded 
}

\DeclareAcronym{VAE-GAN}{
	short = VAE-GAN,
	long  = variational autoencoder-generative adversarial network,
	class = excluded
}

\DeclareAcronym{EVs}{
	short = EVs,
	long  = electric vehicles,
	class = excluded
}

\DeclareAcronym{KL}{
	short = KL,
	long  = Kullback-Leibler,
	class = excluded
}

\DeclareAcronym{EMAmath}{
	short = $\bar{p}_{ema}$,
	long  = {exponential moving average of the real-time price},
	sort  = $\bar{p}_{ema}$,
	class = nomencl
}

\begin{document}
\title{Advancing Generative Artificial Intelligence and Large Language Models for Demand Side Management with Internet of Electric Vehicles}

\author{Hanwen Zhang\raisebox{0.5ex}{\orcidlink{0000-0002-6295-4753}},~\IEEEmembership{Student Member,~IEEE},~Ruichen Zhang\raisebox{0.5ex}{\orcidlink{0000-0002-6859-3645}},~\IEEEmembership{Member,~IEEE,}~Wei Zhang\raisebox{0.5ex}{\orcidlink{0000-0002-2644-2582}},~\IEEEmembership{Member,~IEEE,} \\ ~Dusit Niyato\raisebox{0.5ex}{\orcidlink{0000-0002-7442-7416}}, ~\IEEEmembership{Fellow,~IEEE,}~Yonggang Wen\raisebox{0.5ex}{\orcidlink{0000-0002-2751-5114}}, ~\IEEEmembership{Fellow,~IEEE,}~Chunyan Miao\raisebox{0.5ex}{\orcidlink{0000-0002-0300-3448}}, ~\IEEEmembership{Fellow,~IEEE}
%
% \thanks{This research is supported by the National Research Foundation, Singapore and Infocomm Media Development Authority under its Future Communications Research \& Development Programme (Grant FCP-SIT-TG-2022-007), Defence Science Organisation (DSO) National Laboratories under the AI Singapore Programme (FCP-NTU-RG-2022-010 and FCP-ASTAR-TG-2022-003), Singapore Ministry of Education (MOE) Tier 1 (RG87/22), the NTU Centre for Computational Technologies in Finance (NTU-CCTF), Seitee Pte Ltd, A*STAR under its MTC Programmatic (Award M23L9b0052), MTC Individual Research Grants (IRG) (Award M23M6c0113), the Ministry of Education, Singapore, under the Academic Research Tier 1 Grant (Grant ID: GMS 693), and SIT’s Ignition Grant (STEM) (Grant ID: IG (S) 2/2023 – 792). (\textit{Corresponding author: Wei Zhang})}
%
\thanks{Hanwen Zhang is with both the College of Computing and Data Science, Nanyang Technological University, Singapore 639798, and the Information and Communications Technology Cluster, Singapore Institute of Technology, Singapore 828608 (e-mail: hanwen001@e.ntu.edu.sg and hanwen.zhang@singaporetech.edu.sg)}
\thanks{Wei Zhang is with the Information and Communications Technology Cluster, Singapore Institute of Technology, Singapore 828608 (e-mail: wei.zhang@singaporetech.edu.sg).}
\thanks{Dusit Niyato, Ruichen Zhang, Yonggang Wen, Chunyan Miao are with the College of Computing and Data Science, Nanyang Technological University, Singapore 639798 (e-mail: dniyato@ntu.edu.sg,  ruichen.zhang@ntu.edu.sg, ygwen@ntu.edu.sg, and ASCYMiao@ntu.edu.sg).}
}

\maketitle

\begin{abstract}
The energy optimization and demand side management (DSM) of Internet of Things (IoT)-enabled microgrids are being transformed by generative artificial intelligence, such as large language models (LLMs). This paper explores the integration of LLMs into energy management, and emphasizes their roles in automating the optimization of DSM strategies with Internet of Electric Vehicles (IoEV) as a representative example of the Internet of Vehicles (IoV). We investigate challenges and solutions associated with DSM and explore the new opportunities presented by leveraging LLMs. Then, we propose an innovative solution that enhances LLMs with retrieval-augmented generation for automatic problem formulation, code generation, and customizing optimization. The results demonstrate the effectiveness of our proposed solution in charging scheduling and optimization for electric vehicles, and highlight our solution's significant advancements in energy efficiency and user adaptability. This work shows LLMs' potential in energy optimization of the IoT-enabled microgrids and promotes intelligent DSM solutions. 
\end{abstract}


\section{Introduction}
\label{sec:introduction}
\IEEEPARstart{D}emand side management (DSM), a set of strategies aimed at optimizing energy consumption by aligning energy demand with supply, plays an important role in optimizing energy usage within microgrids. Its importance has grown with the ever-increasing integration of renewable energy sources, \ac{EVs}, distributed \ac{ES}, etc. Implementing DSM strategies is often facilitated by \ac{EMS}, a platform for managing and optimizing energy resources. To achieve the optimal DSM, an EMS must provide accurate forecasting of energy demand and supply and allow optimal resource scheduling. This is typically modeled as an optimization problem where advanced algorithms are of great importance to achieve efficient energy optimization and grid reliability.

DSM optimization has been an established research topic. Early-stage efforts are based on traditional methodologies, such as deterministic approaches, heuristics, and stochastic methods \cite{DSM_Review1}. Recent research advancements for DSM are mostly \ac{ML} based, especially \ac{DAI}. As optimization in the end corresponds to energy-related control actions, \ac{RL}, as a type of ML specialized for system control, has been well studied for DSM \cite{DSM_Review2}. 
A prominent challenge of existing methods is the lack of automation. The DSM optimization cycle involves problem formulation, code generation, and customized optimization, all of which remain largely manual and error-prone. As such, \ac{EMS} developers need to build up domain-specific expertise and evaluate DSM scenarios case by case to improve their decision-making. However, decision imperfections and errors do exist due to the problem's complexity, and DSM optimization can be suboptimal, time-consuming, and, in some cases, introduce system damage and safety violations. \Ac{GenAI} and \ac{LLM} represent the most advanced ML technologies and offer promising solutions for DSM. Existing works show that \ac{EMS} can be significantly enhanced by \ac{LLM} and \ac{GenAI}. However, the current focus is mostly on specific aspects such as prediction \cite{Ref_LLM_for_EV} and anomaly detection \cite{Survey_Ref45} and DSM optimization remains highly expertise-based and non-automatic. 

In this paper, we bridge the gap by discussing \ac{GenAI} and \ac{LLM}'s usage in DSM. We also demonstrate the strength of \ac{GenAI} and \ac{LLM} in automating DSM optimization by proposing a solution based on LLM and \ac{RAG}. In our solution, we first introduce an \textit{Automatic Optimization Formulation} module to automate the problem formulation process, making the selection and application of methods systematic and error-free. Subsequently, we propose the \textit{Automatic Code Generation} module that translates mathematical models into executable code with minimal human intervention, ensuring accurate and efficient implementation. Finally, the end users can further validate and customize this code through an \textit{Automatic Customizing Optimization} module, adapting the solution to specific needs without deep programming expertise. Our solution is designed to be user-friendly and enables both \ac{EMS} developers and end users to solve optimization problems effectively with minimized errors. The main contributions of this paper are as follows. 
\begin{itemize}
    \item We review the recent advancements in energy optimization and DSM, highlighted their limitations, and introduced the use of \ac{GenAI} and \ac{LLM} to advance optimization and management. 
    \item We show that the proposed \ac{LLM} with \ac{RAG} framework can effectively translate end-user objectives and constraints expressed in natural language into a well-posed \ac{MILP} formulation and executable solver code, thereby enabling user-tailored \ac{DSM} schedules without requiring expertise in mathematical programming or software implementation.
    \item We highlight the advantages of the proposed \ac{LLM} and \ac{RAG} based workflow over conventional static \ac{DSM} optimization pipelines, demonstrating that new user preferences, device configurations, and tariff structures can be accommodated by modifying the natural language specification and retrieved knowledge while reusing the underlying \ac{MILP} solver, thus significantly reducing the engineering effort needed to maintain and extend energy management systems. 
    \item Our numerical results further show that, when comparing with a baseline prompt-only \ac{LLM} and the default manufacturer scheduling strategy, the proposed workflow yields better-formulated optimization models and improved DSM performance in terms of operating cost, and user-oriented scheduling metrics. 
\end{itemize}

\section{Background and Related Work}
\label{sect2:Basics}

In this section, we present the background of \ac{DSM} as well as the relevant works of energy-efficient \ac{EMS} and RAG-based \ac{LLM}.

\subsection{Background of Demand Side Management}
The growing importance of sustainable AI solutions in energy management is exemplified by recent initiatives from major cloud providers, such as Google, Microsoft, Amazon Web Services (AWS), and NVIDIA, which invest in highly energy-efficient AI data centers, green cloud computing platforms, and low-power AI accelerators to reduce the carbon footprint of large-scale AI workloads. For example, Google's sustainable data center and AI efficiency initiatives \cite{web_susAI_google}; Microsoft's work on energy-efficient AI and sustainable data centers \cite{web_susAI_microsoft}; AWS sustainable cloud computing and datacenter efficiency \cite{web_susAI_AWS}; and NVIDIA's sustainable computing solutions for AI data centers \cite{web_susAI_NVIDIA}.
Many countries have introduced energy optimization and DSM programs to motivate consumers to adjust their electricity usage patterns and levels, e.g., Singapore \cite{web_SG_EMA} and U.S. \cite{web_US_DepEnergy}. In general, the programs involve two main components: energy efficiency and \ac{DR} \cite{web_DADS_Report}. The former refers to the long-term reduction in energy consumption. This is achieved by adopting advanced techniques or processes that provide the same level of performance or service quality as before but with lower energy use. For example, this could involve installing energy-efficient appliances, upgrading the air conditioning system, or improving optimization algorithms for energy management. The latter is a strategy used in energy systems to adjust electricity demand by consumers in response to supply conditions for a short-term adaptation. It involves, for instance, encouraging consumers to reduce or shift their energy usage during peak periods \cite{springer_RN779}. 

\begin{figure*}[t]
    \centering
    \includegraphics[width=0.98\textwidth]{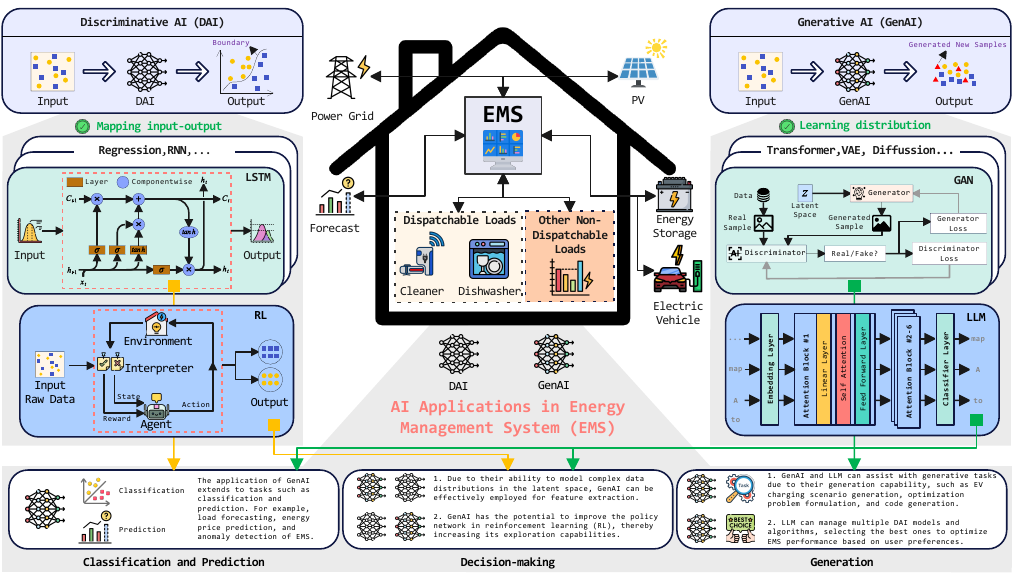}
    \caption{\textbf{Overview of EMS with various AI approaches} --- An illustration of \ac{DAI} and \ac{GenAI} applications within \ac{EMS}, where \ac{EMS} acts as the platform for implementing DSM. The central part highlights a representative \ac{EMS} layout, featuring the forecasting module, renewable energy source, \ac{ES}, \acs{EV}, along with both dispatchable and non-dispatchable loads. Ultimately, \ac{DAI} and \ac{GenAI} collaborate to enable efficient and secure operations of the \ac{EMS}.}
    \label{fig:EMS_AI}
\end{figure*}

As illustrated in \cref{fig:EMS_AI}, the DSM is realized through the \ac{EMS} platform, which is generally integrated with a forecasting module, renewable energy sources, \ac{ES}, \ac{EVs}, as well as both dispatchable and non-dispatchable loads. \Ac{DAI} techniques, such as regression, \ac{RNN}, and \ac{LSTM}, are applicable to the forecasting module \cite{DSM_Ref13,DSM_Ref14r71}, which can include tasks, e.g., predicting electricity prices and loads. On the other hand, \ac{RL} can be utilized within \ac{EMS} for the decision making, e.g., the optimal scheduling of dispatchable loads, EVs, and \ac{ES}, aiming to minimize operating costs \cite{DSM_Ref15}. \ac{GenAI} such as Transformer can be used for load forecasting, \ac{VAE} for anomaly detection, and diffusion models for data augmentation \cite{Survey_Hanwen}. Additionally, emerging \ac{LLM} demonstrates its potential in supporting generative tasks such as problem formulation \cite{Ruichen2} and scenarios generation \cite{Ref_LLM_for_EV}. 

\begin{table*} %[t]
\caption{The Summary of Different Review Papers \\ \textcolor{cyan}{\dmark}: Insight; \textcolor{green}{\cmark}: pros; \textcolor{red}{\xmark}: cons.}
\label{tab:review_paper_summary}
\begin{tblr}{
  width = \linewidth,
  % colspec = {m{0.09\linewidth}m{0.07\linewidth}m{0.1\linewidth}m{0.635\linewidth}}, 
  % colspec = {m{0.09\linewidth}m{0.03\linewidth}m{0.1\linewidth}m{0.675\linewidth}}, 
  colspec = {m{0.09\linewidth}m{0.03\linewidth}m{0.09\linewidth}m{0.685\linewidth}}, 
  cells = {c},     
  hlines,
  vline{2-4} = {-}{},
  vline{2} = {2}{-}{},
  hline{1,8} = {-}{0.08em},
}
\hline
\textbf{Scenario} & 
\textbf{Ref.} & 
\textbf{Method} & 
\textbf{Pros \& Cons} \\
% Scenario 
\SetCell[r=5]{c} Home EMS &   
% Reference    
\cite{DSM_Ref4} & 
% Methodology    
Binary particle swarm (BPS)   &    
% Pros & Cons   
\begin{minipage}[c]{\linewidth}
    \begin{itemize}[noitemsep,topsep=0pt,leftmargin=8pt]
       \item[\textcolor{cyan}{\dmark}] A BPS optimization method for smart home EMS with appliances, renewable energy, ES, and EV.
       \item[\textcolor{green}{\cmark}] Effectively reduce electricity cost and grid dependency across multiple renewable and EV scenarios.
       \item[\textcolor{red}{\xmark}] Need to extend to multi-objective optimization, green hydrogen integration, and microgrid coordination.
    \end{itemize}
\end{minipage} \\
%
% Scenario 
Home EMS &   
% Reference    
\cite{DSM_Ref5} & 
% Methodology    
MILP   &    
% Pros & Cons   
\begin{minipage}[c]{\linewidth}
    \begin{itemize}[noitemsep,topsep=0pt,leftmargin=8pt]
       \item[\textcolor{cyan}{\dmark}] An MILP-based planning framework for off-grid smart homes.
       \item[\textcolor{green}{\cmark}] Achieve about 41\% cost reduction by jointly exploiting flexible demand and vehicle-to-home strategy.
       \item[\textcolor{red}{\xmark}] Need to integrate DSM and V2H into off-grid planning and extend the framework to a larger scale.
    \end{itemize}
\end{minipage} \\
%
% Scenario 
Home EMS &   
% Reference    
\cite{DSM_Ref10} & 
% Methodology    
MILP   &    
% Pros & Cons   
\begin{minipage}[c]{\linewidth}
    \begin{itemize}[noitemsep,topsep=0pt,leftmargin=8pt]
       \item[\textcolor{cyan}{\dmark}] A smart-home framework for scheduling flexible loads, PV, ES, and an EV with grid trading. 
       \item[\textcolor{green}{\cmark}] Deliver 48\% cost reduction and 55\% peak-to-average ratio reduction through joint MILP optimization. 
       \item[\textcolor{red}{\xmark}] Require extension to microgrids coordinating several homes and broader renewable resources. 
    \end{itemize}
\end{minipage} \\
%
% Scenario 
Home EMS &   
% Reference    
\cite{DSM_Ref1} & 
% Methodology    
VAE-GAN   &    
% Pros & Cons   
\begin{minipage}[c]{\linewidth}
    \begin{itemize}[noitemsep,topsep=0pt,leftmargin=8pt]
       \item[\textcolor{cyan}{\dmark}] A VAE–GAN-based generator for high-resolution smart-home load, PV, and EV time series. 
       \item[\textcolor{green}{\cmark}] Outperforms Gaussian mixture model and vanilla GAN in distance metrics and home EMS profit. 
       \item[\textcolor{red}{\xmark}] Reliance on single-home scope and call for scaling to buildings and communities. 
    \end{itemize}
\end{minipage} \\
%
% Scenario 
Home EMS &   
% Reference    
\cite{DSM_Ref16} & 
% Methodology    
Multi-agent DRL   &    
% Pros & Cons   
\begin{minipage}[c]{\linewidth}
    \begin{itemize}[noitemsep,topsep=0pt,leftmargin=8pt]
       \item[\textcolor{cyan}{\dmark}] A hierarchical multi-agent framework for coordinating residential energy use and comfort. 
       \item[\textcolor{green}{\cmark}] Lead to 30.41\% lower energy use and 28.57\% lower peak load with higher comfort and system utility. 
       \item[\textcolor{red}{\xmark}] Suggest incorporating federated learning and blockchain for stronger privacy, security, and trust. 
    \end{itemize}
\end{minipage} \\
%
% Scenario 
\SetCell[r=2]{c} Building thermal system &   %  building thermal comfort
% Reference    
\cite{Ref_RAG1} & 
% Methodology    
RAG-based LLM   &    
% Pros & Cons   
\begin{minipage}[c]{\linewidth}
    \begin{itemize}[noitemsep,topsep=0pt,leftmargin=8pt]
       \item[\textcolor{cyan}{\dmark}] An RAG-based system for unified building thermal comfort monitoring and decision support.
       \item[\textcolor{green}{\cmark}] Yield 94\% accuracy, 92\% precision, and 89\% recall for thermal comfort analysis. 
       \item[\textcolor{red}{\xmark}] Call for larger deployments, stronger fault handling, and voice and augmented reality extensions.
    \end{itemize}
\end{minipage} \\
%
% Scenario 
Building thermal system &   % Central cooling systems
% Reference    
\cite{Ref_RAG2} & 
% Methodology    
RAG-based LLM   &    
% Pros & Cons   
\begin{minipage}[c]{\linewidth}
    \begin{itemize}[noitemsep,topsep=0pt,leftmargin=8pt]
       \item[\textcolor{cyan}{\dmark}] A framework for diagnosis of design, operational, and performance anomalies in cooling systems. 
       \item[\textcolor{green}{\cmark}] Offer 95\% rule-space reduction and 89.08\% diagnostic accuracy via the proposed method. 
       \item[\textcolor{red}{\xmark}] Limited by building diversity, manual knowledge-base updates, and high LLM cost for deployment. 
    \end{itemize}
\end{minipage} \\
%
% Scenario 
Building energy sectors &   
% Reference    
\cite{Ref_RAG3_review} & 
% Methodology    
LLM   &    
% Pros & Cons   
\begin{minipage}[c]{\linewidth}
    \begin{itemize}[noitemsep,topsep=0pt,leftmargin=8pt]
       \item[\textcolor{cyan}{\dmark}] A survey mapping LLM roles in building efficiency and decarbonization.
       \item[\textcolor{green}{\cmark}] Analyze 7 LLM applications across control, code generation, data infrastructure, report analysis, etc.
       \item[\textcolor{red}{\xmark}] Suggest domain-specific LLMs, multimodal approaches, and deeper AI–energy collaboration. 
    \end{itemize}
\end{minipage} \\
%
% Scenario 
\SetCell[r=2]{c} EV charging &   % EV charging system
% Reference    
\cite{Ref_RAG4} & 
% Methodology    
RAG-based LLM   &    
% Pros & Cons   
\begin{minipage}[c]{\linewidth}
    \begin{itemize}[noitemsep,topsep=0pt,leftmargin=8pt]
       \item[\textcolor{cyan}{\dmark}] A multi-agent scheme aligning user choices with grid load via user and station agents and negotiation. 
       \item[\textcolor{green}{\cmark}] Enable dynamic pricing negotiation that enhances load alignment while balancing revenue. 
       \item[\textcolor{red}{\xmark}] Investigate broader data use, long-term behavior, and LLM scale. 
    \end{itemize}
\end{minipage} \\
%
% Scenario 
EV charging &    % EV charging networks
% Reference    
\cite{Ref_RAG5} & 
% Methodology    
RAG-based LLM   &    
% Pros & Cons   
\begin{minipage}[c]{\linewidth}
    \begin{itemize}[noitemsep,topsep=0pt,leftmargin=8pt]
       \item[\textcolor{cyan}{\dmark}] An anomaly detection framework for identifying malicious behavior in EV charging networks.
       \item[\textcolor{green}{\cmark}] Accuracy improvement from 0.65 to 0.85 out of 1.0 with RAG-enhanced LLM classification.
       \item[\textcolor{red}{\xmark}] Lack multi-agent security, adaptive RL defenses, and federated learning for decentralized detection.
    \end{itemize}
\end{minipage} \\
\end{tblr}
\end{table*}

\subsection{Energy-Efficient Energy Management System}
\label{subsec:energy_efficient_EMS}

Recently, energy efficiency has been a central theme in contemporary research \cite{DSM_Ref6,DSM_Ref8,DSM_Ref9,DSM_Ref11}. A growing number of works concentrates on refining energy dispatch strategies so as to improve the overall operational efficiency of the energy management system \cite{DSM_Ref4,DSM_Ref5,DSM_Ref10,DSM_Ref1,DSM_Ref16}. 
For instance, the authors of \cite{DSM_Ref4} formulated a binary linear programming problem to schedule residential appliances by jointly considering user preferences, electrical and thermal loads, renewable energy availability, as well as the constraints of battery and \ac{EV}, and solved it with a binary particle swarm optimization algorithm to obtain near-optimal appliance schedules under real-time pricing. 
In \cite{DSM_Ref5}, the authors presented an \ac{MILP} problem for the optimal planning of off-grid smart home electrification with respect to sizing decisions of \ac{PV} generation, battery storage, diesel backup units, and the operational scheduling of flexible demand and vehicle-to-home (V2H) capabilities. Their goal is to keep the total net present cost of the electrification project as low as possible while also taking into account how investments and operations interact over time. 
In \cite{DSM_Ref10}, the authors developed a multi-objective \ac{MILP} model for a comprehensive home \ac{EMS}. This model optimizes the scheduling of controllable appliances, thermostatically controlled loads, \ac{PV} generation, battery storage, and bidirectional \ac{EV} operations under energy trading. By combining an enhanced $\varepsilon$-constraint method with lexicographic optimization, their method reduced the daily energy cost, peak-to-average ratio, and user discomfort index all at once.
A \ac{VAE-GAN} framework was proposed in \cite{DSM_Ref1} to generate synthetic smart-home datasets for energy management. This model learns joint time-series patterns of household demand, rooftop \ac{PV} generation, and \ac{EV} charging to train an RL-based home \ac{EMS} using synthetic data. By comparing statistical distance measures and the resulting online operating profit, their method shows that the generated datasets better match real distributions and improve economic performance over Gaussian mixture and standard generative models.
The authors of \cite{DSM_Ref16} proposed a hierarchical incentive-driven multi-agent \ac{DRL} framework for smart home \ac{EMS}. The framework coordinates dispatchable and non-dispatchable appliances via the interactions among the system operator, aggregators, and residential users modeled as a Stackelberg game with reputation-based incentives. Their method combines this game-theoretic structure with a multi-agent deterministic policy gradient method. This cuts down on energy use and peak load in homes while making users more comfortable and the system as a whole more useful. 

Although these studies provide rigorous formulations and demonstrate the benefits of advanced scheduling strategies, their optimization pipelines remain largely hand-crafted and tightly coupled to specific scenarios. In practice, \ac{EMS} developers must manually derive complex mathematical models, encode them in solver-specific programming frameworks, and repeatedly adjust both the formulations and the implementation whenever device portfolios, tariff structures, or user preferences change \cite{Ruichen6}. This process is time-consuming and error-prone. It creates a substantial barrier for those who lack deep expertise in mathematical programming and software engineering. As a result, maintaining and extending DSM solutions in IoT-enabled microgrids can be cumbersome, which highlights the need for more systematic and automated frameworks that standardize model construction and reduce the engineering effort required to keep energy-efficient \ac{EMS} deployments up to date.

\subsection{RAG-based LLM}

Research on \ac{GenAI} has accelerated in recent years, with considerable attention directed toward its integration into interactive agent systems \cite{Hongyang1,Ref1_IoTJ,Ref2_IoTJ,Ref3_IoTJ}. 
For example, the authors of \cite{Ref_RAG1} developed an agentic RAG-LLM-based information system that integrates building information modeling, sensor, weather, and guideline data to support thermal comfort monitoring and energy-efficient building management, achieving 94\% accuracy, 92\% precision, and 89\% recall and outperforming a standard RAG-LLM baseline. 
A graph-constrained association rule mining-based energy efficiency diagnosis framework for \ac{HVAC} systems was proposed in \cite{Ref_RAG2}. The authors utilize a knowledge-enhanced LLM with RAG that evaluates candidate rules using design parameters, expert knowledge, and maintenance records, achieving about 89.1\% diagnostic accuracy while reducing the number of candidate rules by more than 95\%. 
At a broader level, the authors of \cite{Ref_RAG3_review} conducted a comprehensive review on the opportunities of applying LLMs in the building energy sector, highlighting potential applications in intelligent control systems, automated code generation, data infrastructure, regulatory compliance, and education for energy efficiency and decarbonization, as well as outlining key challenges and future research directions. 
the authors of \cite{Ref_RAG4} proposed a multi-agent-LLM-based \ac{EV} charging system in which user and EV charging station agents, empowered by LLMs with RAG capabilities and conditional generative adversarial network generated data, collaboratively provide personalized charging recommendations and dynamic price adjustments, thereby balancing user preferences, charging efficiency, and grid load stability under power dispatching requirements. 
Further within the EV ecosystem, the authors of \cite{Ref_RAG5} proposed an LLM-powered agentic anomaly detection system for \ac{EV} charging networks, where RAG-enhanced \ac{LLM} analyze real-time charging session data and domain-specific cyber-security knowledge to detect attacks such as billing fraud, energy theft, and overcharging, achieving higher accuracy and fewer false positives than a baseline \ac{LSTM} autoencoder.

However, these aforementioned studies overall primarily deploy \ac{LLM} and \ac{RAG} as auxiliary tools for monitoring, diagnosis, recommendation, or anomaly detection within existing energy and \ac{EV} management workflows, or discuss potential applications and challenges of \ac{LLM} in the energy sector at a conceptual level, rather than using \ac{GenAI} as an engine that explicitly constructs and updates the underlying \ac{DSM} optimization models. In particular, they do not address an IoT-enabled microgrid in which the \ac{EMS} must jointly schedule distributed energy resources, flexible loads, and \ac{EV} charging through a formal mixed-integer optimization problem, nor do they leverage \ac{LLM} with \ac{RAG} to translate natural-language specifications from developers and end users into consistent \ac{MILP} formulations and solver-ready code. In contrast, this paper positions an \ac{RAG}-enhanced \ac{LLM} as a modeling and code-generation engine that retrieves \ac{DSM}-related technical documents and executable templates as external knowledge, synthesizes well-posed \ac{MILP} models and optimization programs from high-level objectives and constraints, and regenerates customized \ac{DSM} schedules as user preferences, device configurations, and tariff structures evolve, thereby complementing conventional static \ac{DSM} optimization pipelines with an adaptive, user-tailored workflow. \Cref{tab:review_paper_summary} provides a summary of the reviewed paper in this section.

\section{ System Model and Demand Side Management}
\subsection{Demand Side Management Formulation}
\label{sect:optimization_problem}

The optimization problem for the demand side management system in a microgrid, such as a house, has been formulated as an \ac{MILP} problem. The microgrid consists of $n_{sto}^{ES}$ \ac{ES} units, $n_{sto}^{EV}$ \ac{EVs}, a \ac{PV} system, the dispatchable loads (e.g., washing machine, cleaner, oven, and dishwasher), and non-dispatchable loads (e.g., refrigerator, security system, network devices, and medical devices). The objective is to minimize the operating cost of the microgrid. This mathematical formulation and optimization will be implemented in the \ac{LLMs} with \ac{RAG} framework \cite{Ruichen2}. \Cref{tab:symbolsOpt} lists symbols used in \cref{eqn:ES1,eqn:ES2,eqn:ES3,eqn:ES4,eqn:ES5,eqn:ES6,eqn:typyX_1,eqn:typyX_2,eqn:type2_3,eqn:type2_4,eqn:type2_5,eqn:PowerBalance,eqn:GridCon1,eqn:GridCon2,eqn:ObjectiveFunction} of \cref{sect:optimization_problem}. The RAG dataset is constructed with the optimization problem including the mathematical formulations of constraints and objective function, as well as the Python code to solve such an optimization problem. 

\begin{table}
	\centering
    \caption{Symbols of ES, EV, and Dispatchable Loads} 
    \label{tab:symbolsOpt}
    % \begin{tabular}{m{2.58cm}||m{5.35cm}}
    \begin{tabular}{m{2.1cm}||m{5.83cm}}
		\toprule
		\Xhline{1pt} 
		\multicolumn{1}{c||}{\textbf{Symbols}} & \multicolumn{1}{c}{\textbf{Description}}  \\ 
		\Xhline{0.75pt} 
		\hline
\multicolumn{2}{c}{Decision Variables} \\ \hline  
            $\lambda_k$
            & Binary status: Type 1 or Type 2 loads \\ \hline
            $\mu^{ch}_{sto_{j}}$,
            & Binary charging status of ES or $EV_{i}$ \\ \hline  % (1: charging, 0: not charging) 
            $\mu^{disch}_{sto_{j}}$
            & Binary discharging status of ES or $EV_{i}$  \\ \hline % (1: discharging, 0: not discharging)
			$\nu_k^s$
            & Binary status: Start operation of Type 2 loads  \\ \hline
            $\nu_k^e$
            &  Binary status: End operation of Type 2 loads \\ \hline
            $P^{ch}_{sto_{j}}$
            & Charge power of the ES or $EV_{i}$  \\ \hline
			$P^{disch}_{sto_{j}}$
            & Discharge power of the ES or $EV_{i}$ \\ \hline
            $P^{imp}_{grid}$, $P^{exp}_{grid}$
            & Power bought from or sold to grid \\ \hline
\multicolumn{2}{c}{Parameters} \\ \hline  
        	$C^{imp}_{grid}$, $R^{exp}_{grid}$
        	& Electricity buy or sell price \\ \hline
        	$E^{rated}_{sto_{j}}$
        	& Nominal capacity of ES or $EV_{i}$ \\ \hline
        	$E^{min}_{sto_{j}}$
        	& Minimum capacity SoC of ES or $EV_{i}$ \\ \hline
        	$E^{max}_{sto_{j}}$
        	& Maximum capacity SoC of ES or $EV_{i}$ \\ \hline
        	$E^{req}_{sto_{j}}$
        	& Required SoC of ES or $EV_{i}$ at $t^{req}_{sto_{j}}$ \\ \hline
        	$E_k$
        	& Total energy required for load $k$ \\ \hline
        	$H_k$
        	& Total duration of load $k$ must operate \\ \hline
        	$\eta^{ch}_{sto_{j}}$
        	& Charge efficiency of the ES or $EV_{i}$   \\ \hline
        	$\eta^{disch}_{sto_{j}}$
        	& Discharge efficiency of the ES or $EV_{i}$ \\ \hline
        	$\mu_{sto_{j}}$
        	& Binary availability status of ES or $EV_{i}$ \\ \hline
        	$P^{ch}_{sto,max}$
        	& Maximum charge power of the ES or $EV_{i}$ \\ \hline
        	$P^{disch}_{sto,max}$
        	& Maximum discharge power of the ES or $EV_{i}$ \\ \hline
        	$P^{imax}_{grid}$, $P^{emax}_{grid}$
        	% & Maximum grid import or export power \\ \hline
            & Maximum power bought from or sold to the grid \\ \hline
        	$P_{\text{Type1}, k}$, $P_{\text{Type2}, k}$
        	& Power consumed by Type 1 or Type 2 load $k$ \\ \hline
        	$P_{PV}$, $P_{load}$
        	& Solar PV Power or non-dispatchable loads  \\ \hline
        	$t$, $\Delta t$, $T$
        	& Time index, time interval or total time horizon  \\ \hline
        	$t^{req}_{sto_{j}}$
        	& Time for ES or $EV_{i}$ reach $E^{req}_{sto_{j}}$  \\ \hline
        	$\mathcal{L}_1$, $\mathcal{L}_2$
        	& Set of all Type 1 or Type 2 loads \\ \hline
\multicolumn{2}{c}{Other Notations} \\ \hline 
            $E_{sto_{j}}$
            & SoC of the ES or $EV_{i}$ \\ \hline
		\Xhline{1pt} 
		\bottomrule
	\end{tabular}
\end{table}

\subsubsection{Objective Function}

The objective function is to minimize the operating cost of the microgrid, i.e., reducing the net purchased power from the grid as shown in \cref{eqn:ObjectiveFunction} \cite{DSM_Ref1,DSM_Ref4}. 

\begin{equation} 
\label{eqn:ObjectiveFunction}
    \min \ \sum_{t=1}^T \left( P^{imp}_{grid}(t) \cdot C^{imp}_{grid}(t) - P^{exp}_{grid}(t) \cdot R^{exp}_{grid}(t) \right) \cdot \Delta t
\end{equation}
where $P^{imp}_{grid}(t) \cdot C^{imp}_{grid}(t)$ is the total cost of purchasing power from the grid. $P^{exp}_{grid}(t) \cdot R^{exp}_{grid}(t)$ is the total revenue from selling power to the grid.

\subsubsection{Constraints}

\paragraph{Storage Devices}
The storage devices consists of \ac{ES} and \ac{EVs} in the system. 
The constraints of \ac{ES} and \ac{EV} are formulated in \cref{eqn:ES1,eqn:ES2,eqn:ES3,eqn:ES4,eqn:ES5,eqn:ES6}. Since \ac{ES} and \ac{EV} are both storage devices, the subscript symbol $sto_{j}$ is defined to represent $ES$ or $EV_{i}$, e.g., $sto_{1}$, $sto_{2}$, $sto_{3}$ represent for $ES$, $EV_{1}$, and $EV_{2}$ respectively. 
\Cref{eqn:ES1} updates the \ac{SoC} of the storage devices in the charging/discharging process \cite{DSM_Ref1,DSM_Ref4,DSM_Ref5,DSM_Ref6,DSM_Ref8,DSM_Ref9,DSM_Ref10,DSM_Ref11}. The \ac{SoC} is updated at time $t+1$ over the time step $\Delta t$, with respect to the nominal capacity of the storage devices, $E^{rated}_{sto_{j}}$, as follows: 
\begin{equation}  \label{eqn:ES1}
    \begin{aligned} 
        E_{sto_{j}}&(t+1) = E_{sto_{j}}(t) - \\
        &\left( \frac{P^{disch}_{sto_{j}}(t)}{\eta^{disch}_{sto_{j}}}-\eta^{ch}_{sto_{j}}\times P^{ch}_{sto_{j}}(t)\right)  \times \frac{100 \times \Delta t}{E^{rated}_{sto_{j}}}
    \end{aligned}
\end{equation}

\Cref{eqn:ES2} and \cref{eqn:ES3} impose constraints on the charging and discharging power of the storage devices, respectively \cite{DSM_Ref4}. Thus, the charging and discharging operations are restricted by their respective maximum/minimum limits and the availability of the storage devices. 
\begin{equation} 	\label{eqn:ES2}
	0 \leq P^{ch}_{sto_{j}}(t) \leq P^{ch}_{ES,max} \times \mu_{sto_{j}}(t) \times \mu^{ch}_{sto_{j}}(t)
\end{equation}
\begin{equation} 	\label{eqn:ES3}
	0\leq P^{disch}_{sto_{j}}(t) \leq P^{disch}_{ES,max} \times \mu_{sto_{j}}(t) \times \mu^{disch}_{sto_{j}}(t)
\end{equation}
where $\mu^{ch}_{sto{j}}(t)$ and $\mu^{disch}_{sto{j}}(t)$ denote the binary indicators that characterize whether the storage device is engaged in charging or discharging at time~$t$. In particular, $\mu^{ch}_{sto{j}}(t)=1$ indicates that the device is in charging operation, whereas $\mu^{disch}_{sto{j}}(t)=1$ reflects that it is performing discharging activity. Meanwhile, $\mu_{sto_{j}}(t)$ represents the binary availability status of the storage device, such that $\mu_{sto_{j}}(t)=1$ signifies that the device is connected to the system and operational, and vice versa.

\Cref{eqn:ES4} ensures that the storage device cannot charge and discharge at the same time \cite{DSM_Ref6}. 
\begin{equation} 	\label{eqn:ES4}
	\mu^{ch}_{sto_{j}}(t) + \mu^{disch}_{sto_{j}}(t) \leq 1  
\end{equation}

\Cref{eqn:ES5} defines the \ac{SoC} of the storage devices at time $t$ can only operate in the range between the minimum and maximum \ac{SoC} required by the storage devices \cite{DSM_Ref4}. 
\begin{equation} 	\label{eqn:ES5}
	 E^{min}_{sto_{j}} \leq E_{sto_{j}}(t) \leq E^{max}_{sto_{j}}
\end{equation}

\Cref{eqn:ES6} ensures that the storage device should maintain a minimum required \ac{SoC}, $E^{req}_{sto_{j}}$, at the specific time, $t^{req}_{sto_{j}}$ \cite{DSM_Ref4,DSM_Ref5}. For example, if $E^{req}_{sto_{j}} = 80\%$, the \ac{SoC} of the storage device should be charged/discharged to maintain at least 80\% at time $t^{req}_{sto_{j}}$. 
\begin{equation} 	\label{eqn:ES6}
    E_{sto_{j}}(t^{req}_{sto_{j}}) \geq E^{req}_{sto_{j}}
\end{equation}

\paragraph{Dispatchable Loads Constraints}
The dispatchable loads are categorized into two types. Type 1 loads are interruptible, meaning the devices can be temporarily interrupted while operating. Type 2 loads are uninterruptible, meaning the devices cannot be interrupted during operation. Each dispatchable load must meet its operational constraints. 

For both Type 1 and Type 2 loads, they should follow the total energy consumption and the total duration constraints in \cref{eqn:typyX_1,eqn:typyX_2} \cite{DSM_Ref9}. The total energy consumption constraint over the scheduling horizon: 
\begin{equation} \label{eqn:typyX_1}
\begin{split}
	& \sum_{t=1}^T P_{\text{TypeX}, k}(t) \cdot \lambda_k(t) \cdot \Delta t = E_k, \\
    & P_{\text{TypeX}, k}(t)=
    \left
    \{\begin{matrix}
    P_{\text{Type1}, k}(t), \qquad \text{if} \quad \forall k \in \mathcal{L}_1
     \\
    P_{\text{Type2}, k}(t), \qquad \text{if} \quad \forall k \in \mathcal{L}_2
    \end{matrix}
    \right.
\end{split}
\end{equation}
where the notation $\forall k \in \mathcal{L}_1$ designates the the $k$-th load associated with the Type 1 loads, whereas $\forall k \in \mathcal{L}_2$ indicates those corresponding to the Type 2 loads.

The total duration constraint: 
\begin{equation} \label{eqn:typyX_2}
	\sum_{t=1}^T \lambda_k(t) = H_k, \quad \forall k \in \mathcal{L}_1 \, \text{or} \, \mathcal{L}_2
\end{equation}

\textbf{Type 1 Loads} refer to interruptible and flexible appliances whose operation can be scheduled over multiple nonconsecutive intervals within the planning horizon \cite{DSM_Ref9}. These loads do not require a continuous operating cycle; instead, the optimizer may divide their operation into several disjoint segments as long as their total energy or total operating duration requirement is satisfied. For example, a robotic vacuum cleaner is modeled as a Type 1 load. The cleaner may be instructed to operate for a short period, pause when the electricity price is high or when occupants are present, and then resume later. \cref{eqn:typyX_1} shows that the scheduled operation must fulfill the total energy requirement of Type 1 loads. For instance, if a cleaner operating power is 0.6 $kW$ and need to operate 4 hours, the total energy required for this load is $0.6 \times 4 = 2.4 \ kWh$. \cref{eqn:typyX_2} imposes the non-consecutive operational time constraint, e.g., the cleaner must operate 4 hours even if it is interrupted. 

\textbf{Type2 Loads} refer to the non-interruptible appliances that must operate in a consecutive manner, i.e., once switched on, the appliance must run continuously for its entire predefined operating cycle without any pauses or gaps between time intervals \cite{DSM_Ref9}. The only scheduling flexibility lies in the choice of when this continuous operating block begins. The user is typically indifferent to the exact time at which the appliance runs, as long as the full cycle is completed before a specified deadline.  
For example, a dishwasher may require a $90$-minute washing cycle that must be executed as one uninterrupted block. The optimizer can schedule this $90$-minute block at different feasible periods within the planning horizon, such as from 22:00 to 23:30 or from 02:00 to 03:30, but it cannot split the cycle into two separate segments, e.g., 22:00–22:30 and 02:00–03:00. Therefore, dishwashers are modeled as Type 2 loads: their entire cycle must run continuously once started, while the start time of this consecutive operating block remains fully dispatchable within the allowed activation window.  
Similarly, \cref{eqn:typyX_1,eqn:typyX_2} provide the total energy and time constraints for the scheduled Type 2 Loads. 
The constraints \cref{eqn:type2_3,eqn:type2_4,eqn:type2_5} ensure that the load operates continuously without interruptions once started, with the appliance being started and stopped only once during the operation \cite{DSM_Ref5,DSM_Ref10,DSM_Ref11}. Specifically, to ensure each Type 2 load $k$ has exactly one start time, the following constraint is imposed:  
\begin{equation} \label{eqn:type2_3}
\setlength{\jot}{-1.5ex}  % Reduces the space between lines
    \begin{aligned}
        \sum_{i=0}^{H_k - 1} \lambda_k(t+i) \geq H_k \cdot \nu^s_k(t), \quad & \forall k \in \mathcal{L}_2, \\
        &\forall t \in \{0, \dots, T - H_k\} 
    \end{aligned}
\end{equation}
where the binary variable $\nu_k^s(t)$ indicates whether the load starts at time $t$. The sum of all these start events must equal 1, i.e., the load starts only once during the scheduling period. \cref{eqn:type2_3} ensures that when the Type 2 load, e.g., the dishwasher starts at time \( t \) (i.e., \( \nu^s_k(t) = 1 \)), it operates continuously for \( H_k \) time steps. 

The linking start and end markers constraint: 
\begin{equation} \label{eqn:type2_4}
    \begin{aligned}
        \nu^e_k(t + H_k - 1) = \nu^s_k(t), \quad & \forall k \in \mathcal{L}_2, \\ 
        &\forall t \in \{0, \dots, T - H_k\} 
    \end{aligned}
\end{equation}
\Cref{eqn:type2_4} links the start and end markers to ensure consistency. If the Type 2 load (e.g., dishwasher) starts at time $t$, it must end at $t + H_k - 1$. It ensures that each Type 2 load $k$ ends exactly once. The binary variable $\nu_k^e(t)$ indicates whether the load ends at time $t$. 

The operation activation consistency: 
\begin{equation} \label{eqn:type2_5}
\setlength{\jot}{-1.5ex}  % Reduces the space between lines
    \begin{aligned}
        \lambda_k(t) \leq \sum_{j = \max(0, t - H_k + 1)}^{t} \nu^s_k(j), \quad & \forall k \in \mathcal{L}_2, \\ 
        & \forall t \in \{0, \dots, T - 1\}
    \end{aligned}
\end{equation}
Constraint in \cref{eqn:type2_5} ensures that $\lambda_k(t)$ (indicating the Type 2 load, e.g., dishwasher is operating) is only active if there is a valid start marker ($\nu^s_k$) in the preceding $H_k$ time steps. It models the tracking of consecutive operations, and ensures that if the load is operating at time $t$ (i.e., $\lambda_k(t) = 1$), it must continue operating at $t + 1$, and similarly, when it stops, the operation will terminate at $t$. The binary variables $\nu_k^s(t)$ and $\nu_k^e(t)$ handle the transitions: if $\nu_k^s(t) = 1$, the load starts at $t$; if $\nu_k^e(t) = 1$, the load ends at $t$.

\paragraph{Power Balance}
The total power consumption (e.g., loads and charging power of \ac{ES} and \ac{EVs}) at each time step must equal the sum of power generated by \ac{PV}, power purchased from grid, and power discharged from \ac{ES} and \ac{EVs}. The power balance constraint is shown in \cref{eqn:PowerBalance} \cite{DSM_Ref1,DSM_Ref4,DSM_Ref5,DSM_Ref11}. 

\begin{equation} \label{eqn:PowerBalance}
\begin{aligned}
    & \underset{\textcolor{blue}{\text{PV}}}{\underbrace{P_{PV}(t)}} + \underset{\textcolor{blue}{\text{Storage Devices}}}{\underbrace{\sum_{j=1}^{n_{sto}} \left ( P^{disch}_{{sto_{j}}}(t) - P^{ch}_{{sto_{j}}}(t) \right )}} + \underset{\textcolor{blue}{\text{Grid}}}{\underbrace{\left ( P^{imp}_{grid}(t) - P^{exp}_{grid}(t) \right )}} \\
    & = \underset{\textcolor{blue}{\text{Loads}}}{\underbrace{ P_{load}(t)}} + \underset{\textcolor{blue}{\text{Type 1 and Type 2 Loads}}}{\underbrace{ \left (\sum_{k \in \mathcal{L}_1} P_{\text{Type1}, k}(t) + \sum_{k \in \mathcal{L}_2} P_{\text{Type2}, k}(t) \right ) \cdot \lambda_k(t)}}
\end{aligned}
\end{equation}
where $P_{PV}(t)$ is the power generated by PV, the term $\sum_{j=1}^{n_{sto}} \left ( P^{disch}_{{sto_{j}}}(t) - P^{ch}_{{sto_{j}}}(t) \right )$ is the net power supplied from the storage devices to the microgrid. $n_{sto}$ refers to the number of storage devices in the system. The term $\left ( P^{imp}_{grid}(t) - P^{exp}_{grid}(t) \right )$ is the net power supplied from the grid. $P_{load}(t)$ are the non-dispatchable loads. The last term $\left (\sum_{k \in \mathcal{L}_1} P_{\text{Type1}, k}(t) + \sum_{k \in \mathcal{L}_2} P_{\text{Type2}, k}(t) \right ) \cdot \lambda_k(t)$ is the sum of Type 1 and Type 2 dispatchable loads.

\paragraph{Grid Constraints}
The power purchased from the grid, $ P^{imp}_{grid}(t)$, and the power sold to the grid, $P^{exp}_{grid}(t)$, are constrained at each time period \cite{DSM_Ref4}.
\begin{equation} \label{eqn:GridCon1}
 0 \leq P^{imp}_{grid}(t) \leq P^{imax}_{grid}(t)
\end{equation}

\begin{equation} \label{eqn:GridCon2}
 0 \leq P^{exp}_{grid}(t) \leq P^{emax}_{grid}(t)
\end{equation}

\section{Proposed Framework}

From the microgrid operator's perspective, the DSM process or lifecycle has two main phases: prediction and optimization, as shown in \cref{fig:EMS_AI}. In the first phase, the prediction involves forecasting variables such as renewable energy generation, electricity prices \cite{Survey_Ref28}, and load demand \cite{Ref1_Load_Forcasting}. Next, the optimization phase aims to enhance \ac{EMS} strategies by determining optimal dispatch schedules for flexible loads, as well as optimizing the charging and discharging processes of \ac{ES} and \ac{EVs}. For these reasons, a full-lifecycle digital twin would be an ideal solution to enable seamless automation across all phases. In recent studies, e.g., \cite{Survey_Ref45}, the role of \ac{GenAI} in enhancing the prediction phases of DSM was demonstrated. However, the research gap remains in automating the optimization phase, which this section seeks to address by exploring automated DSM optimization techniques. Hence, we introduce a \ac{GenAI} agent to address key challenges in \ac{EMS} development for IoT-enabled microgrid (\cref{fig:outline}), such as the complexity of optimization problem formulation and the difficulty of coding implementation. 

\subsection{General Framework}

\begin{figure}[t]
    \centering
    \includegraphics[width=1.0\columnwidth]{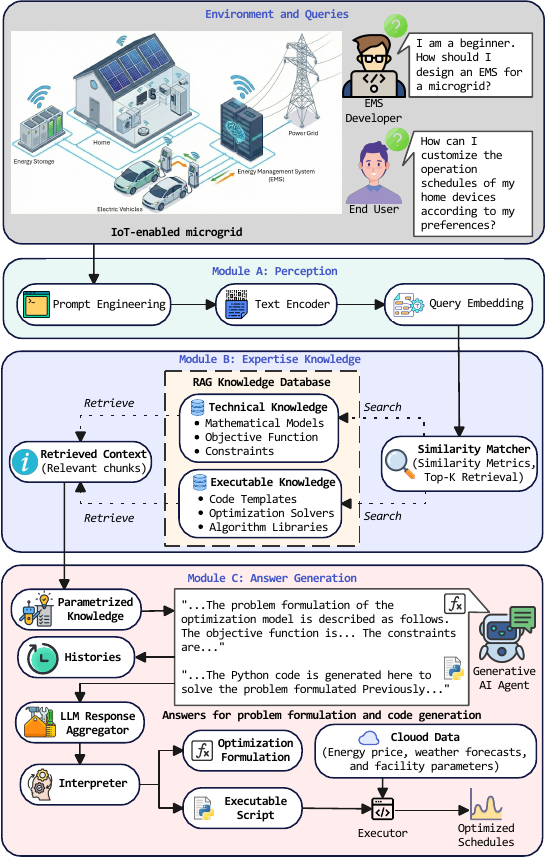}
    \caption{A general framework design of a GenAI agent supporting modeling and code generation for IoT-enabled microgrid. Specifically, to address the challenges faced by EMS developers for IoT-enabled microgrid—such as complex optimization problem formulation and the difficulty of coding implementation—and to enable non-experts (e.g., end users) to customize EMS scripts, we introduce a generative AI agent that supports these tasks through natural language interaction. The generative AI agent is responsible for formulating optimization problems, generating coding scripts, and modifying existing scripts according to the requirements of both EMS developers and end users.}
    \label{fig:outline}
\end{figure}

A general framework is shown in \cref{fig:outline}, demonstrating how the \ac{GenAI} agent assists an EMS developer and end user in addressing IoT-enabled microgrid-related queries. Initially, an IoT-enabled microgrid query is raised by the EMS developer or the end user. 
This query then goes through the \textbf{Module A: Perception} which consists of prompt engineering, text encoder, and query embedding \cite{Ruichen2}. The prompt engineering organizes task context, constraints, and objectives in the input prompt to guide the \ac{LLM} toward producing accurate and task-relevant outputs. The text encoder converts input text into semantic vector representations that allow the \ac{LLM} to understand and process linguistic meaning. It provides the resulting query embedding which serves as a unified semantic representation that enables the subsequent retrieval module to match the user query against relevant technical and executable knowledge stored in the RAG database. 

In \textbf{Module B: Expertise Knowledge}, we employ a similarity matcher to measure the semantic relevance between the query embedding and the vectorized entries in the \ac{RAG} knowledge database \cite{Ruichen6,Ref_RAG_Framework1}. It uses a similarity metric to extract the most pertinent knowledge segments that are semantically aligned with the user query. The \ac{RAG} knowledge database is composed of two complementary components: \textit{technical knowledge} and \textit{executable knowledge}. The first knowledge consists of domain-specific expertise, such as the mathematical models, objective functions, and system constraints that formally describe the optimization problems in IoT-enabled microgrid \cite{DSM_Ref6,DSM_Ref8,DSM_Ref9,DSM_Ref11,DSM_Ref4,DSM_Ref5,DSM_Ref10,DSM_Ref1,DSM_Ref16}. The second knowledge contains the solver-ready code templates, optimization solvers, and algorithm libraries that support the direct implementation of the formulated optimization problems \cite{web_Pyomo_solver,web_Pyomo_package}. Using similarity-based retrieval, the most relevant technical and executable knowledge is chosen and given to the downstream generation module as context for retrieval. 

In \textbf{Module C: Answer Generation}, the retrieved context from \textbf{Module B} is integrated with the LLM’s intrinsic parameterized knowledge and the original user query to guide the generative inference process \cite{Ruichen2,Ruichen6}. Then, each response from the LLM is recorded in the histories and collected by the LLM response aggregator in \cref{fig:outline}. The interpreter receives the aggregated information and presents the optimization formulation in a user-friendly manner, together with an executable script for further processing. The interpreter, implemented as a deterministic post-processing layer, receives the aggregated multi-turn LLM outputs, extracts and structures the key artifacts for the queried task, renders the finalized optimization formulation for human inspection (e.g., by compiling the generated LaTeX equations), and assembles the returned code fragments into an executable script that can be downloaded by the user or dispatched to the executor for solving. Upon preparation of the script, the executor executes it utilizing the most recent energy price forecasts, weather predictions, and facility parameters obtained from the cloud to provide optimized operational schedules for devices within the IoT-enabled microgrid. In this case, the executor is a runtime component that 1) acquires and verifies the recent cloud-derived inputs, 2) executes the Python script generated by the interpreter, and 3) transforms the solver output into device-specific schedules \cite{web_TSP1,web_TSP2}. 

\subsection{Module A: Perception}

A local corpus $\mathcal{D} = \{ d_1, \dots, d_N \}$ is built from DSM-related documents, which are subsequently segmented into chunks. Each chunk $d_j$ is then encoded by a text embedding model through a standard embedding interface, which maps input text into a $d$-dimensional dense vector space for semantic retrieval and similarity search \cite{Yinqiu_DataCenter,Ruichen2}. In our implementation, we instantiate this interface using LangChain (e.g., \texttt{OpenAIEmbeddings}) with the OpenAI \texttt{text-embedding-ada-002} model (with $d=1536$) \cite{web_openAI_embedding}, while other embedding providers can be substituted without changing the overall RAG pipeline \cite{web_langchain_doc1}. 
Let $q$ denote a user query expressed in natural language, and $\mathcal{D}$ represent a set of preprocessed text chunks. Since $f_{\mathrm{emb}}(\cdot)$ denotes the embedding function which is used for both queries and chunks, they are projected into the same $1536$-dimensional embedding space \cite{web_openAI_embedding}. The embedding vectors of the query and the $j$-th chunk are given by:
\begin{equation}
    \boldsymbol{u}(q)=f_{\mathrm{emb}}(q), \qquad
    \boldsymbol{v}_j=f_{\mathrm{emb}}(d_j), \quad j=1,\dots,N,
    \label{eq:embeddings}
\end{equation}
where $\boldsymbol{u}(q),\boldsymbol{v}_j \in \mathbb{R}^{dim}$ denote the dense vector representations of query and the $j$-th chunk respectively where $dim = 1536$.

\subsection{Module B: Expertise Knowledge}

\subsubsection{Similarity}
To measure how relevant a chunk $d_j$ is to the query $q$, we evaluate their distance directly in the embedding space and define the similarity score as the negative squared Euclidean distance \cite{RAG_eqn_ref3,RAG_eqn_ref4}: 
\begin{equation}
    \mathrm{s}(q,d_j)
    = - \big\|\boldsymbol{u}(q) - \boldsymbol{v}_j\big\|_2^2
    = - \sum_{k=1}^{\mathrm{dim}} \big(u_k(q) - v_{j,k}\big)^2,
    \label{eq:similarity}
\end{equation}
where $u_k(q)$ and $v_{j,k}$ are the $k$-th components of $\boldsymbol{u}(q)$ and $\boldsymbol{v}_j$ respectively. A larger value of $\mathrm{s}(q,d_j)$ corresponds to a smaller Euclidean distance between $\boldsymbol{u}(q)$ and $\boldsymbol{v}_j$, meaning that $d_j$ is more similar to $q$ in the embedding space. 
The embedding model maps text into a semantic vector space where semantically related texts appear close to each other. The similarity score $\mathrm{s}(q,d_j)$ therefore measures how close the query and the chunk are in this semantic space, enabling the retriever to select the most relevant pieces of knowledge.

\subsubsection{RAG Knowledge Database}
\label{subsubsec:RAG_Database}

Mathematical programming based \ac{DSM} for IoT-enabled microgrid is challenging since an \ac{EMS} must jointly model distributed energy resources, flexible appliances, grid interactions, tariff dynamics, and user-driven requirements in a consistent optimization program \cite{DSM_Ref5,DSM_Ref10}. When customizing an \ac{EMS} optimization model for a given microgrid, the first step is to establish a specialized database. This is due to the fact that  there is no off-the-shelf corpus that simultaneously provides rigorous \ac{DSM} formulations and solver-ready implementations that can be reliably reused across different device portfolios and preference settings. To support subsequent automatic optimization formulation, code generation, and customization, we construct an \ac{RAG} database that integrates two complementary types of knowledge. Specifically, the database contains technical knowledge, including the formal system model with the objective function in \cref{eqn:ObjectiveFunction} and constraints in \cref{eqn:ES1,eqn:ES2,eqn:ES3,eqn:ES4,eqn:ES5,eqn:ES6,eqn:typyX_1,eqn:typyX_2,eqn:type2_3,eqn:type2_4,eqn:type2_5,eqn:PowerBalance,eqn:GridCon1,eqn:GridCon2} of an IoT-enabled microgrid, as well as executable knowledge, such as code snippets and solver-ready templates. In addition, the database is augmented with time-series profiles (e.g., electricity prices and renewable generation) and facility parameters retrieved from the cloud, enabling the RAG-based LLM to retrieve task-relevant expertise and implementation references for downstream generation and customization.

\subsubsection{Retriever Mechanism}
With the similarity scores in \cref{eq:similarity} defined for every query--chunk pair $(q,d_j)$, the retriever implements a deterministic top-$K$ nearest-neighbour search in the embedding space \cite{Ref_RAG_Framework1}. For a given query $q$, the collection of similarity values $\{ s(q,d_1), s(q,d_2), \dots, s(q,d_N) \}$ is first computed between the query embedding $\boldsymbol{u}(q)$ and all chunk embeddings $\{ \boldsymbol{v}_j \}_{j=1}^{N}$ as specified in \cref{eq:embeddings}. For a retrieval budget $K \le N$, we define the index set of the $K$ most relevant chunks as \cite{RAG_eqn_ref2}: 
\begin{equation}
    \mathcal{I}_{K}(q)
    =
    \operatorname*{TopK}_{j \in \{1,\dots,N\}}
    s(q,d_j),
    \label{eq:topk}
\end{equation}
where $\operatorname*{TopK}$ returns the indices corresponding to the $K$ largest values in the collection $\{ s(q,d_j) \}_{j=1}^N$. The set of retrieved documents is therefore \cite{RAG_eqn_ref5}:
\begin{equation}
    \mathcal{R}(q)
    =
    \{\, d_j \mid j \in \mathcal{I}_K(q) \,\}.
\end{equation}

For clarity, the retriever can be regarded as a deterministic operator mapping a query to its retrieved context set \cite{RAG_eqn_ref2}:
\begin{equation}
    \mathcal{R}(q)
    =
    \operatorname{Ret}(q; \mathcal{D})
    :=
    \big\{ d_j \in \mathcal{D} \,\big|\, j \in \mathcal{I}_K(q) \big\}.
    \label{eq:retr_mech}
\end{equation}
In our case study, the retrieval module is configured to return the top $K=3$ most relevant chunks for every query. Accordingly, $\operatorname{Ret}(q;\mathcal{D})$ (or equivalently $\mathcal{R}(q)$) consists of the three elements in $\mathcal{D}$ whose similarity scores $s(q,d_j)$ attain the highest values, thereby providing the most relevant context for subsequent generation.

\subsection{Module C: Answer Generation}

The retrieved chunks in $\mathcal{R}(q)$ are concatenated with the original query $q$ by a deterministic aggregation function $\mathrm{Agg}(\cdot,\cdot)$ to form the final textual context $c(q)$ which is the textual context obtained by concatenating the query and the top-$K$ retrieved chunks \cite{RAG_eqn_ref1,RAG_eqn_ref2}:
\begin{equation}
    c(q) = \mathrm{Agg}\big(q, \mathcal{R}(q)\big),
    \label{eq:context}
\end{equation}
$c(q)$ is then provided as input to the LLM in the subsequent generation stage. The aggregation operator,  $\mathrm{Agg}(\cdot,\cdot)$, is instantiated by the retrieval layer of the LLM-based framework, which constructs a single textual prompt by concatenating the user query with the top-$K$ retrieved chunks and inserting the resulting string into the model context as part of the conversational history. The overall behavior of this retrieval and prompt-construction procedure can therefore be abstracted by the operator $\mathrm{Agg}(\cdot,\cdot)$ in \cref{eq:context}.

Let $\boldsymbol{y} = (y_1,\dots,y_T)$ denote the output sequence generated by the LLM, where $y_t$ is the $t$-th token and $T$ is the query-dependent sequence length determined by the decoding procedure. Conditioned on the aggregated context $c(q)$, the retrieval-augmented model defines the following conditional distribution \cite{RAG_eqn_ref2,RAG_eqn_ref6}:
\begin{equation}
    p_{\theta}^{\mathrm{RAG}}(\boldsymbol{y} \mid q)
    :=
    p_{\theta}\big(\boldsymbol{y} \mid c(q)\big),
    \label{eq:rag_deterministic}
\end{equation}
where $p_{\theta}$ denotes the autoregressive language model parameterized by $\theta$. $p_{\theta}^{\mathrm{RAG}}(\boldsymbol{y} \mid q)$ is the conditional probability that the fixed-parameter LLM assigns to the token sequence $\boldsymbol{y}$ when it is prompted with the retrieval-augmented context $c(q)$. In this formulation, the retrieval-augmented model first forms the context $c(q)$ by concatenating the query with the elements of $\mathcal{R}(q)$ and then generates the answer sequence $\boldsymbol{y}$ according to $p_{\theta}(\boldsymbol{y} \mid c(q))$, so that all retrieved information influences generation exclusively through its inclusion in the initial textual context.

In this paper, the LLM is used only in inference mode: its parameters $\theta$ are fixed and provided by an external service, and no additional training or fine-tuning is carried out within our framework. Here, a generative pre-trained transformer (GPT) is employed to model $p_{\theta}(\boldsymbol{y} \mid c)$, leveraging retrieved knowledge to produce contextually grounded and coherent responses concerning DSM with EV modeling. Following the standard formulation of decoder-only LLMs, we model the LLM as follows. For any given textual context $c$, an autoregressive conditional distribution over the output token sequence $\boldsymbol{y} = (y_1,\dots,y_T)$ \cite{RAG_eqn_ref2,RAG_eqn_ref7,RAG_eqn_ref8}:
\begin{equation}
    p_{\theta}(\boldsymbol{y} \mid c)
    =
    \prod_{t=1}^{T}
    p_{\theta}\big(y_t \mid y_{<t}, c\big),
    \label{eq:llm_autoreg}
\end{equation}
where $y_t$ denotes the $t$-th output token, $y_{<t} = (y_1,\dots,y_{t-1})$ is the prefix generated so far, and $T$ is the query-dependent sequence length determined by the decoding procedure.

\subsection{Summary of General Framework}
To summarize, the proposed agent follows an end-to-end pipeline that transforms a natural-language microgrid query into solver-ready artifacts and optimized operating schedules (\cref{fig:outline}). In \textbf{Module A}, the user query $q$ is organized via prompt engineering and encoded into an embedding $\boldsymbol{u}(q)$ using an embedding function $f_{\mathrm{emb}}(\cdot)$ as in \cref{eq:embeddings}. In \textbf{Module B}, the retriever evaluates semantic relevance in the embedding space (e.g., via $\mathrm{s}(q,d_j)$ in \cref{eq:similarity}) and returns the top-$K$ context chunks $\mathcal{R}(q)$ according to \cref{eq:retr_mech}. In \textbf{Module C}, the retrieved context is aggregated with the original query through $c(q)=\mathrm{Agg}\big(q,\mathcal{R}(q)\big)$ in \cref{eq:context} and fed to the LLM to produce task-specific outputs under the autoregressive generation model in \cref{eq:llm_autoreg}. The multi-turn LLM outputs are then consolidated by the response aggregator, after which the interpreter structures the results into human-readable optimization formulations (e.g., rendered equations) and assembles the generated code into an executable script. Finally, the executor binds this script with the latest cloud-sourced inputs (e.g., price forecasts, weather predictions, and facility parameters) and runs the corresponding solver stack to output device-level optimized schedules for the IoT-enabled microgrid.

\subsection{An Example Workflow of Proposed Framework}

\begin{figure*}[t]
    \centering
    \includegraphics[width=0.98\textwidth]{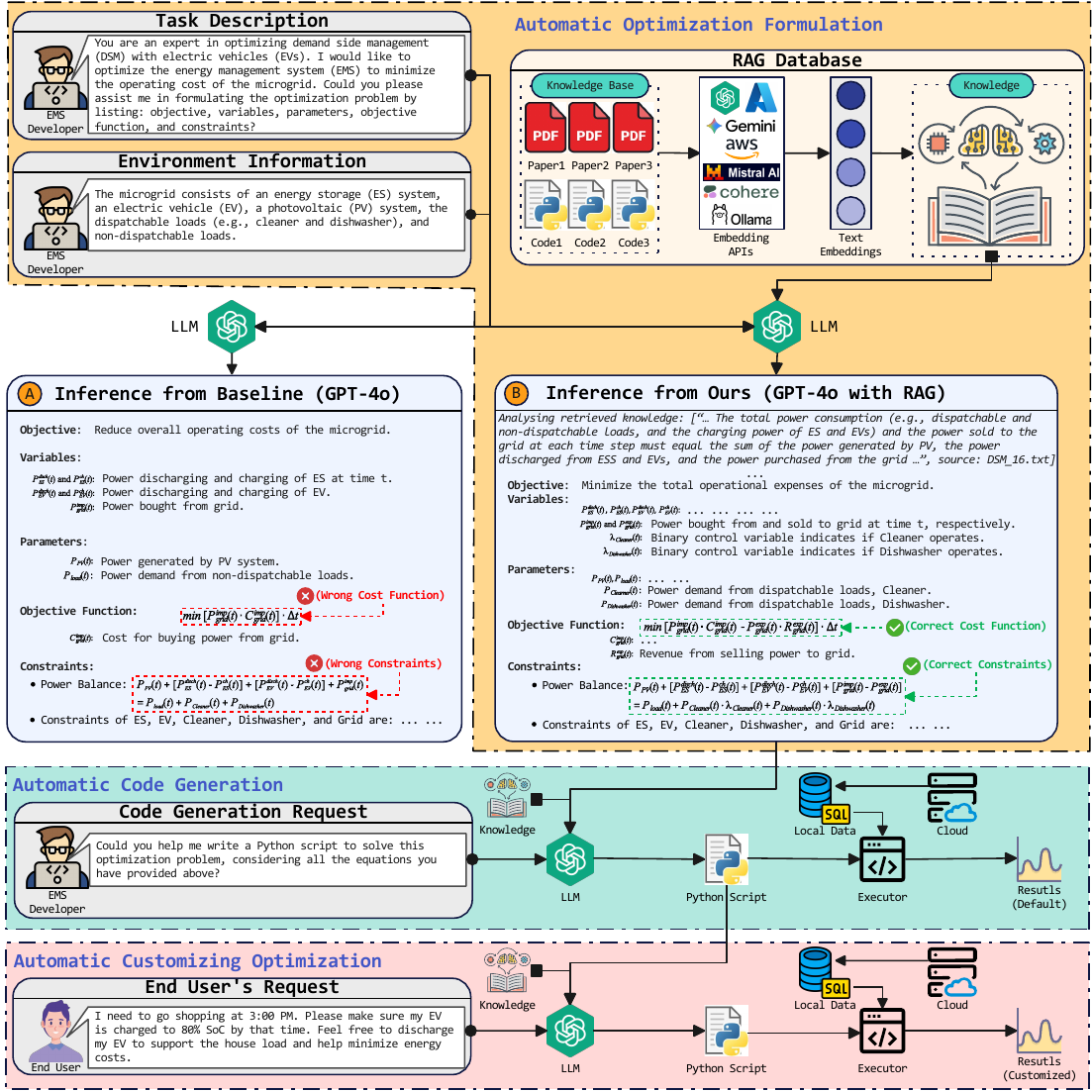}
    \caption{\textbf{An overview of the function-oriented workflow under the proposed general framework.} --- RAG-based LLM for energy optimization and DSM: Automatic optimization formulation, code generation, result generation based on user's request. A \& B: The comparison of the optimization formulations between the backbone GPT-4o and the proposed RAG-enhanced LLM.}
    \label{fig:LLM_code_generation}
\end{figure*}

The DSM is implemented through the \ac{EMS}, which enables an IoT-enabled microgrid to achieve cost-effective energy usage, reduced peak demand, and improved resource utilization. Building upon the general framework in \cref{fig:outline}, we implement a function-oriented workflow for LLM-enabled DSM that automates the optimization process for both the \ac{EMS} developer and the end user. Specifically, this workflow converts a natural-language DSM query into a solver-ready optimization formulation, executable program code, and tailored operational outcomes through an RAG-based LLM. Hence, it supports the three core stages proposed in this paper—\textit{Automatic Optimization Formulation}, \textit{Automatic Code Generation}, and \textit{Automatic Customizing Optimization}—as illustrated in \cref{fig:LLM_code_generation}.

\subsubsection{Stage 1: Automatic Optimization Formulation}

The EMS developer's primary responsibility is to create optimization algorithms that effectively manage energy consumption within the microgrid. This procedure requires acquiring a wide range of characteristics and data, including energy usage, changing pricing models, and demand patterns. These elements are then interconnected using complex mathematical models. Traditionally, this endeavor has been characterized by its considerable labor demands and dependence on the developer's expertise, a skill set that necessitates substantial time to master and may not fully capture the dynamic aspects of energy systems, including demand shifts and regulatory adaptations. To mitigate these difficulties, we introduce an \textit{Automatic Optimization Formulation} module, as shown in \cref{fig:LLM_code_generation}, which utilizes an RAG-based LLM \cite{Ref_IoTM1}. We design this module to automate the formulation of optimization problems, dynamically adjusting to the specific characteristics of the IoT-enabled microgrid; for instance, a smart home EMS incorporating EVs, ES, and household appliances. The automated formulation process serves to diminish the need for manual intervention and specialized domain knowledge, thereby streamlining the modeling procedure and facilitating the development of precise and scalable solutions for the optimal operation of the microgrid. 

The proposed RAG-based LLM for energy optimization and DSM is shown in \cref{fig:LLM_code_generation}. In this case, the RAG database is supplied with technical and executable knowledge constructed following the design in \cref{subsubsec:RAG_Database}, thereby enhancing LLM's capability in DSM tasks such as peak shaving, load shifting, and energy storage optimization. The developer's inputs, which include the task description, environmental information, optimization goals, and key factors, are converted into vectors and then provided to the \acs{LLM}. Following this, the RAG-based LLM that is specialized for DSM computes the semantic distances between the inputs and all stored chunks. The most relevant chunks are selected as external knowledge, which the \acs{LLM} uses to generate responses that are accurate and coherent and effectively support the optimization problem formulation. For example, the optimization problem formulated by \ac{LLM} with \ac{RAG} is shown in the block \textbf{\textit{B - Inference from Ours (GPT-4o with RAG)}} of \Cref{fig:LLM_code_generation}.

\subsubsection{Stage 2: Automatic Code Generation}
Following the establishment of the optimization problem, the following task involves translating these mathematical formulations into executable code. This stage often requires certain programming skills or experiences to ensure the accurate and effective execution of the optimization models. However, developing code manually is both time-consuming and costly, and it is prone to errors. To automate this complex process, we propose the \textit{Automatic Code Generation} module in \cref{fig:LLM_code_generation} which automates the complex process by generating correct, high-quality, executable code from the optimization model. This allows EMS developers to apply optimization solutions faster, reduce errors, improve performance, and scale to customized requirements with less manual programming. 

In \Cref{fig:LLM_code_generation}, the \textit{Automatic Code Generation} module is illustrated to solve the formulated optimization problem, enabling the \ac{EMS} developer to submit code generation requests through prompts, such as requesting the writing of a script to solve the previously formulated optimization problem. Then, the \acs{LLM} generates a Python script based on prior responses from the automatic optimization formulation module, taking into account the \ac{EMS} developer's prompts and retrieving relevant knowledge from the \ac{RAG} database. The executor first extracts the time-series data—such as predicted energy prices and forecasted renewable energy generation—along with facility parameters from the cloud. Subsequently, the executor runs the generated script and provides the default optimization results from the manufacturer. 

\subsubsection{Stage 3: Automatic Customizing Optimization Stage}
For the end user, the key requirement is the ability to adapt the optimization results to their specific needs in real time. While automatic optimization formulation and code generation streamline the \ac{EMS} development process, they must also be flexible enough to respond to end-user requests. For instance, an end user may wish to prioritize the operation of specific appliances or adjust the scheduling of \ac{ES}, \ac{EVs}, and consumption based on their unique operational goals or preferences. To meet this need, we propose \textit{Automatic Customizing Optimization}. This module dynamically revises optimization results based on user inputs. With an RAG-based LLM, the system is able to interpret user requests, modify the optimization models accordingly, and generate updated schedules or results in real time. Hence, the \ac{EMS} become responsive, user-centric, and aligned with both individual preferences and overall system efficiency. By reducing the need for manual intervention, end users gain the ability to make timely, optimized, and informed decisions that improve performance and resource utilization. 

For example in \cref{fig:LLM_code_generation}, the end user can request customization of optimal schedules for facilities such as an \acs{EV}. The \acs{LLM} then analyzes this request, retrieves the relevant \ac{RAG} knowledge, and modifies the manufacturer's provided code to generate a new and customized optimization code automatically. The resulting customized optimal schedules can be achieved by executing this newly generated code.

\section{Results and Discussion}
\label{section:case_study}
In this section, we validate the proposed framework for automating energy optimization and DSM with LLM.

\begin{table}[t]
\centering
\caption{Parameters of facilities in Microgrid}
\label{tab:Parameters}
\begin{tblr}{width = 1.0\linewidth,
  colspec = {Q[1.8,c,m] Q[1.5,c,m] Q[5,l,m]},
  % colspec = {Q[1.8,c,m] Q[1.5,c,m] Q[4.7,l,m]},
  row{1} = {c}, 
  hlines,
  vline{2-3} = {-}{}, 
  vline{2} = {2}{-}{},
}
\hline
\textbf{Parameters} & \textbf{Values} & \textbf{Description} \\
\SetCell[c=3]{c} System Parameters \\
%
% Parameters
$\Delta t$ &
% Values
30 minutes   &
% Description
Time step to update the data  \\
\SetCell[c=3]{c} Energy Storage (ES) \\
% Parameters
Efficiency&
% Values
0.95   &
% Description
Efficiency of charging/discharging ES  \\
%
% Parameters
Capacity &
% Values
12 kWh   &
% Description
Rated capacity of ES  \\
%
% Parameters
Min SoC &
% Values
20\%   &
% Description
Min SoC limits of ES  \\
%
% Parameters
Max SoC &
% Values
100\%   &
% Description
Max SoC limits of ES  \\
%
% Parameters
Target SoC &
% Values
100\%   &
% Description
Min required SoC at target time  \\
%
% Parameters
Target Time &
% Values
23:00   &
% Description
Time to reach target SoC  \\
%
% Parameters
Charge &
% Values
4 kW   &
% Description
Max charging power of ES  \\
%
% Parameters
Discharge &
% Values
4 kW   &
% Description
Max discharging power of ES  \\
\SetCell[c=3]{c} Electric Vehicle (EV) \\
%
% Parameters
Efficiency &
% Values
0.95   &
% Description
Efficiency of charging/discharging EV  \\
%
% Parameters
Capacity &
% Values
40 kWh   &
% Description
Rated capacity of EV  \\
%
% Parameters
Min SoC &
% Values
20\%   &
% Description
Min SoC limits of EV  \\
%
% Parameters
Max SoC &
% Values
80\%   &
% Description
Max SoC limits of EV  \\
%
% Parameters
Target SoC &
% Values
100\%   &
% Description
Min required SoC at target time  \\
%
% Parameters
Target Time &
% Values
7:30   & % 15   &
% Description
Time to reach target SoC  \\
%
% Parameters
Charge &
% Values
7 kW   &
% Description
Max charging power of EV  \\
%
% Parameters
Discharge &
% Values
7 kW   &
% Description
Max discharging power of EV  \\
\SetCell[c=3]{c} Dispatchable Load --- Cleaner (Type 1) \\
%
% Parameters
Power &
% Values
0.6 kW   &
% Description
Rated charging power of cleaner \\
%
% Parameters
Duration &
% Values
4 hours   &
% Description
Total charging hours of cleaner  \\
\SetCell[c=3]{c} Dispatchable Load --- Dishwasher (Type 2) \\
%
% Parameters
Power &
% Values
1 kW   &
% Description
Rated power of Dishwasher \\
%
% Parameters
Duration &
% Values
1 hour   &
% Description
Total operating hours of Dishwasher  \\
\hline
\end{tblr}
\end{table}

\subsection{Setup and Time-series Dataset}

To assess the effectiveness of the proposed RAG-based LLM solution for DSM, we begin by using it to model a representative DSM optimization problem for EV charging, specifically the optimal scheduling problem. The \ac{LLM} with the \ac{RAG} framework, as proposed in \cite{Ruichen2}, is used, along with a tutorial on its application \cite{RAG_tutorial_code}. We present a microgrid scenario that includes a home \ac{EMS} connected to an \ac{ES}, an \acs{EV}, a \ac{PV}, a cleaner, a dishwasher, and other non-dispatchable loads in \cref{fig:EMS_AI}. The time-series dataset of \ac{PV} power generation and non-dispatchable residential loads was collected from the Customer Led Network Revolution (CLNR) project in the UK \cite{Dataset1_TC5}. The energy price data were obtained from the provider, Energy Stats UK \cite{Dataset2_price}. The parameters of facilities including \ac{ES}, \acs{EV}, cleaner, and dishwasher are listed in \cref{tab:Parameters}. The dispatchable loads could be categorized into two types. Type 1 loads are interruptible, meaning the devices can be temporarily interrupted while operating, e.g., cleaner. Type 2 loads are uninterruptible, meaning the devices cannot be interrupted during operation, e.g., dishwasher.

\subsection{Evaluation of Problem Formulation}
\label{sect:Evaluation}

To validate the problem formulation, we provide the same prompt which outlines the objective of minimizing the operating cost of the microgrid, to both the baseline GPT-4o and our GPT-4o with \ac{RAG} the model. As illustrated in the block \textbf{\textit{A - Inference from Baseline (GPT-4o)}} of \Cref{fig:LLM_code_generation}, the output of the baseline GPT-4o is general and superficial. It successfully formulates the objective function to minimize the total operating cost associated with purchasing power from the grid, along with the power balance constraint. However, it overlooks several factors that could impact optimization performance. These include the absence of revenue from selling power to the grid in the objective function and the failure to incorporate the variables (e.g., $\lambda_{Cleaner}(t)$ and $\lambda_{Dishwasher}(t)$) for controlling the dispatchable loads of the cleaner and dishwasher in the power balance equation. In contrast, our RAG-based LLM as shown in block \textbf{\textit{B - Inference from Ours (GPT-4o with RAG)}} of \Cref{fig:LLM_code_generation} uses a semantic router to direct queries to the most relevant documents. It then retrieves the essential expertise, integrates it with the pre-trained knowledge of \ac{LLM}, and generates the final problem formulation accurately. Notably, both the revenue from selling power to the grid in the objective function and dispatchable loads in the power balance constraint are taken into account in our solution.

\begin{figure}[t]
    \centering
    \includegraphics[width=1.0\columnwidth]{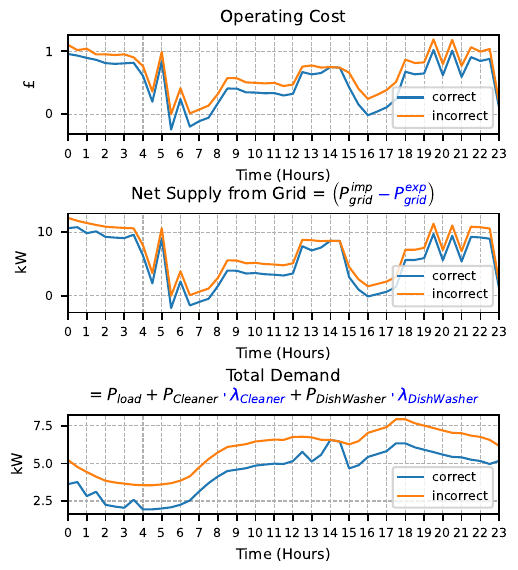}
    \caption{\textbf{Incorrect vs. correct problem formulations} --- A numerical comparison between the incorrect and correct optimization models is presented, where the incorrect formulation omits the variable representing power sold to the grid, $P^{\mathrm{exp}}_{\mathrm{grid}}$, as well as the control variables for dispatchable loads, namely $\lambda_{\mathrm{Cleaner}}(t)$ and $\lambda_{\mathrm{DishWasher}}(t)$, which govern the on/off operation of household devices.}
    \label{fig:compare_correct_incorrect}
\end{figure}

Specifically, consider a scenario where the system is allowed to sell electricity back to the grid, and the selling price is assumed to be equal to the purchasing price. The baseline model (GPT-4O) takes EMS developer's task description and environment information but incorrectly formulates this scenario by omitting the power-selling capability in the objective function. Moreover, the constraint set does not include control variables for dispatchable loads, such as the cleaner and dishwasher. Consequently, the relevant constraints in \cref{eqn:typyX_1,eqn:typyX_2,eqn:type2_3,eqn:type2_4,eqn:type2_5} are not enforced, which leads to these dispatchable loads operating without control. This small misalignment can result in a substantial increase in the overall operating cost of the IoT-enabled microgrid. For example, \cref{fig:compare_correct_incorrect} presents a numerical comparison between the incorrectly formulated model produced by block \textit{\textbf{A – Inference from Baseline (GPT-4o)}} and the correctly formulated model generated by block \textit{\textbf{B – Inference from Ours (GPT-4o with RAG)}}. 

In the top subplot of \cref{fig:compare_correct_incorrect}, the incorrect model increases operating costs over time by omitting the revenue term of selling power to the grid (optimizing $\sum_{t=1}^T P^{imp}_{grid}(t) \cdot C^{imp}_{grid}(t) \cdot \Delta t$ instead of \cref{eqn:ObjectiveFunction}), removing economic incentive and decision flexibility to offset costly grid purchases via selling energy or scheduling actions such as ES or EV use. 
The middle subplot of \cref{fig:compare_correct_incorrect} shows a larger net supply from the grid due to the baseline (GPT4o) formulation, which excludes the selling decision $P^{exp}_{grid}(t)$, causing $P^{imp}_{grid}(t)-P^{exp}_{grid}(t)$ to degenerate to $P^{imp}_{grid}(t)$, increasing reliance on buying power from grid when local resources could be balanced through bidirectional exchange.
The incorrect model in the bottom subplot of \cref{fig:compare_correct_incorrect} inflates total demand due to improper dispatchable-load controllability encoding. The baseline power-balance expression adds dispatchable load powers without the binary activation variables, such as $\lambda_{\mathrm{Cleaner}}(t)$ and $\lambda_{\mathrm{DishWasher}}(t)$, which are missing from the total demand equation: $P_{load}(t)+P_{\mathrm{Cleaner}}(t)\lambda_{\mathrm{Cleaner}}(t)+P_{\mathrm{DishWasher}}(t)\lambda_{\mathrm{DishWasher}}(t)$. 
The three subplots of \cref{fig:compare_correct_incorrect} show that the incorrect formulation overestimates demand, generates higher grid supply, and raises operating cost, while the correct formulation preserves dispatchable-load decisions and grid selling economics for lower-cost, price-responsive scheduling.

\subsection{Evaluation of Code Generation}

\begin{figure}[t]
    \centering
    \includegraphics[width=1.0\columnwidth]{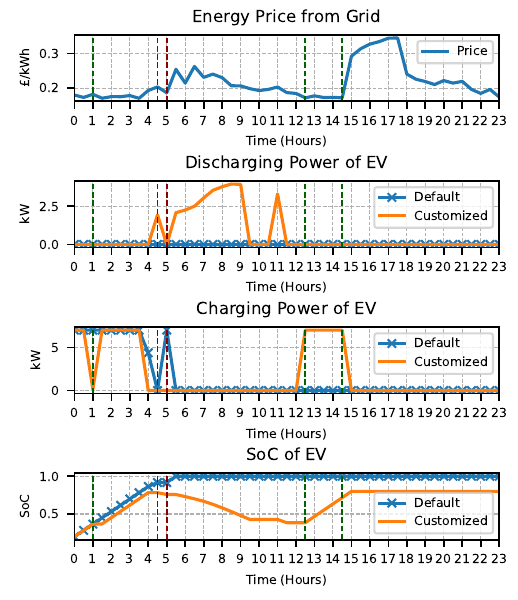}
    \caption{\textbf{Simulation results of customizing optimization} --- Comparison of the default optimization from the manufacturer and the customized optimization from the end user. Default: Optimized result from the manufacturer. Customized: Optimized result from the end user.}
    \label{fig:Ch_EV_SoC}
\end{figure}

Subsequently, the effectiveness of the \textit{Automatic Optimization Formulation}, the \textit{Automatic Code Generation}, and  the \textit{Automatic Customizing Optimization} modules are examined through the execution of two Python scripts that were generated by the \acs{EMS} developer and the end user with our proposed solution (GPT-4o with RAG) respectively. 
\Cref{fig:Ch_EV_SoC} presents a comparison between the scheduling results of the \acs{EV} charging plan under the default manufacturer configuration and the customized configuration set by the end user. Specifically, the user requests that the EV's \ac{SoC} must be at least 80\% by 3 PM. In default method, the system is configured to charge the EV to 100\% SoC by 7:30 am, with an initial SoC of 20\%. The system prioritizes charging during the hours of 0:00 am to 4:00 am when energy prices are lowest and stops charging briefly at 4:30 am due to a slight increase in energy price, resuming at 5:00 am when prices drop again. The EV ceases charging entirely at 5:30 am upon reaching the required 100\% SoC. Additionally, the default method does not permit the EV to discharge power, even during periods of high energy prices, thus missing opportunities to reduce operational costs. In contrast, the customized method optimizes charging and discharging schedules based on user preferences, which request the EV to reach 80\% SoC by 3:00 pm and allow it to discharge power to support the house load during high-price periods. The system strategically halts charging at 1:00 am, considering the relatively higher energy price compared to the midday period (12:30 pm to 2:30 pm). It resumes charging during these midday hours to achieve the desired 80\% SoC by 3:00 pm. Additionally, the customized method enables the EV to discharge at 4:30 am, from 5:30 am to 9:00 am, and at 11:00 am, effectively offsetting the house load during high-price periods and minimizing the microgrid's operational costs. 

The default method incurs an operating cost of \pounds 25.33, whereas the customized method reduces this to \pounds 23.27,reflecting an 8.13\% cost reduction. Our approach demonstrates a more cost-effective and user-tailored energy management strategy compared to the default method. Unlike traditional optimization methods, the LLM with RAG framework in \cref{fig:LLM_code_generation} improves adaptability and generalization by dynamically generating optimization formulations from user input and context, without requiring retraining or coding expertise. 

To further clarify how these scheduling behaviors and cost differences arise from the proposed framework, we next relate the observed results to the underlying optimization reformulation and workflow of an RAG-based LLM. In particular, the transition from the manufacturer default schedule to the user-customized schedule is triggered by a natural-language preference update (``$\mathrm{SoC}\geq 80\%$ by 15:00'' and permitting discharge), which is processed by the perception and retrieval modules to fetch the corresponding \ac{DSM} constraint templates and solver-ready implementation snippets from the \ac{RAG} database. Conditioned on the retrieved technical and executable knowledge, the generation module reformulates the underlying \ac{MILP} (i.e., updating the time-indexed \ac{SoC} requirement and the charge/discharge decision logic) and regenerates a consistent Python solver script without manual re-derivation or re-coding by the user. The executor then binds this regenerated script with the same cloud-sourced time-series inputs (prices, \ac{PV}, and non-dispatchable loads) and computes the new optimal schedule, yielding the price-responsive charging/discharging patterns and the cost reduction observed in \Cref{fig:Ch_EV_SoC}. Therefore, \Cref{fig:Ch_EV_SoC} serves as an empirical verification that the proposed RAG-enhanced LLM can operationalize user intent into solver-consistent model and code updates. It enables rapid customization and maintaining optimization correctness as preferences and operational constraints evolve. 

\section{Conclusions}
\label{sect:Conclusions}
In this paper, we have investigated a relationship between ML and DSM. It begins by reviewing the research efforts with traditional approaches, GenAI, and LLM for energy optimization and DSM, along with discussions about different technologies' strengths and limitations. Then, we have demonstrated the potential of GenAI and LLM for DSM with a case study, where the development of an RAG-enhanced \acs{LLM} solution has been designed and developed to automate DSM optimization. The results obtained by our automated solution have shown that DSM can be significantly enhanced with automated optimization and intelligent DR with EV charging/discharging for energy supply-demand balancing and cost minimization. Our research in this paper offers valuable insights for advancing energy optimization and DSM with the latest ML advancements and inspires future studies.

%%%%%%%%% References %%%%%%%%% 
\balance
\bibliographystyle{IEEEtran}
\bibliography{Refbib}
\newpage

\end{document}